\documentclass[letterpaper]{article} 
\pdfoutput=1
\usepackage{aaai23}  
\usepackage{times}  
\usepackage{helvet}  
\usepackage{courier}  
\usepackage[hyphens]{url}  
\usepackage{graphicx} 
\usepackage{natbib}  
\usepackage{caption} 
\usepackage{algorithm}
\usepackage{algorithmic}

\usepackage{newfloat}
\usepackage{listings}
\DeclareCaptionStyle{ruled}{labelfont=normalfont,labelsep=colon,strut=off} 
\floatstyle{ruled}
\newfloat{listing}{tb}{lst}{}
\floatname{listing}{Listing}
\usepackage{amsmath,amssymb} 
\usepackage{xcolor}
\usepackage{adjustbox}
\usepackage{multirow}
\usepackage{graphicx}
\usepackage{subcaption}
\usepackage{booktabs}
\usepackage{tabularx}
\usepackage{enumitem}

\newcommand{\lrvq}{$\text{LR}^2\text{VQ}$}
\newcommand{\lra}{$\mathbf{A}$}
\newcommand{\lrb}{$\mathbf{B}$}
\newcommand{\lrwp}{$\mathbf{W}^{'}$}
\newcommand{\lrwhp}{$\mathbf{\widehat{W}}^{'}$}
\newcommand{\lrw}{$\mathbf{W}$}
\newcommand{\lrd}{$\tilde{d}$}
\newcommand{\cb}{$\boldsymbol{C}$}
\newcommand{\cd}{$\boldsymbol{I}$}
\newcommand{\ah}{$\mathbf{\widehat{A}}$}
\newcommand{\lrwr}{$\mathbf{W}_r$}
\title{Learning Low-Rank Representations for Model Compression}
\author {
    Zezhou Zhu,\textsuperscript{\rm 1}
    Yucong Zhou, \textsuperscript{\rm 2}
    Zhao Zhong \textsuperscript{\rm 2}
}
\affiliations {
    \textsuperscript{\rm 1} Beijing University of Posts and Telecommunications\\
    \textsuperscript{\rm 2} Huawei\\
    zhuzezhou@bupt.edu.com, {zhouyucong1, zorro.zhongzhao}@huawei.com
}

\usepackage{bibentry}

\begin{document}
\nocopyright

\maketitle

\begin{abstract}
Vector Quantization (VQ) is an appealing model compression method to obtain a tiny model with less accuracy loss.
While methods to obtain better codebooks and codes under fixed clustering dimensionality have been extensively studied, optimizations of the vectors in favour of clustering performance are not carefully considered, especially via the reduction of vector dimensionality.
This paper reports our recent progress on the combination of dimensionality compression and vector quantization, proposing a Low-Rank Representation Vector Quantization (\lrvq) method that outperforms previous VQ algorithms in various tasks and architectures.
\lrvq\ joins low-rank representation with subvector clustering to construct a new kind of building block that is directly optimized through end-to-end training over the task loss.
Our proposed design pattern introduces three hyper-parameters, the number of clusters $k$, the size of subvectors $m$ and the clustering dimensionality \lrd.
In our method, the compression ratio could be directly controlled by $m$, and the final accuracy is solely determined by \lrd.
We recognize \lrd\ as a trade-off between low-rank approximation error and clustering error and carry out both theoretical analysis and experimental observations that empower the estimation of the proper \lrd\ before fine-tunning.
With a proper \lrd, we evaluate $\text{LR}^2\text{VQ}$ with ResNet-18/ResNet-50 on ImageNet classification datasets, achieving 2.8\%/1.0\% top-1 accuracy improvements over the current state-of-the-art VQ-based compression algorithms with 43$\times$/31$\times$ compression factor.
\end{abstract}


\section{Introduction}
In recent years, deep neural networks have achieved remarkable performance on different vision tasks like image classification, object detection, and semantic segmentation.
However, this progress is fueled by wider and deeper network architectures, which possess a large memory footprint and computation overhead.
These networks are hard to be deployed on battery-powered and resource-constrained hardware such as wearable devices and mobile phones.
At the same time, deep neural networks are redundant in parameters and layer connections, which implies that it is possible to compress them without much accuracy loss.
Therefore, compressing neural networks is essential for real-world applications.

There are several approaches to compressing deep neural networks from different perspectives.
Compact network design\cite{zhang2017shufflenet,sandler2018mobilenetv2,howard2019searching,tan2019efficientnet} and neural architecture search (NAS)\cite{zoph2018learning,zhong2018blockqnn,pham2018efficient,DBLP:conf/nips/ZhangZZ20,DBLP:journals/corr/abs-1912-11191} focus on exploring lightweight architectures;
Pruning\cite{NIPS1989_6c9882bb} tends to eliminate unnecessary connections or channels on heavy models directly;
Quantization\cite{courbariaux2016binaryconnect,hubara2016quantized} compresses each parameter from FP32 to lower bits.
Here we concentrate on Vector Quantization(VQ)\cite{10.1109/TPAMI.2010.57} to achieve a higher compression-to-accuracy ratio.

VQ exploits filter redundancy to reduce storage cost with \emph{codebook} and \emph{codes}.
Most of the prior works\cite{gong2014compressing,son2018clustering,wu2016quantized,stock2020bit,Chen2020TowardsCN,martinez2021permute} demonstrate impressive performance on compression-to-accuracy ratio.
The key to the success of VQ is clustering and fine-tuning, where clustering decreases the memory demand for storage and computation, and fine-tuning helps recover the model performance by data-driven optimization.
The accuracy loss of the quantized network is closely related to the clustering error.
Great progress have been achieved on finding more expressive codebooks and codes by reducing the clustering error under a \textit{fixed} clustering dimensionality.
However, the clustering dimensionality is a critical factor in impacting clustering error.
Applying a distance-based clustering method in high-dimensional space is more likely to suffer from the curse of dimensionality\cite{article}, which introduces significant difficulties for clustering to produce large clustering errors.
On the contrary, clustering in low-dimensional space is much easier to produce lower clustering error.
Specifically, the dimensionality of subvectors in convolutional layers is always 9 or 18, raising the difficulty in finding expressive codebooks and codes in high dimensional space, and resulting in poor quantization performance.
On the question of dimensionality redunction, low-rank representation methods are widely used for all sorts of dimensionality reduction tasks.
These remarkable methods invoke the demand to investigate the possibility to design significantly better quantizers by the mixture of VQ and low-rank representations.

To keep up with the above demand, this paper proposes Low-Rank Representation Vector Quantization (\lrvq), a new method that jointly considers the compression ratio and the clustering dimensionality to achieve better quantization.
\lrvq\ utilizes low-rank representations (LRRs) to approximate the original filters and clustering the subvectors in LRRs to achieve quantization.
Our method possesses an extraordinary characteristic: the compressed model size is controlled by $m$ (the subvector size), and the quantization performance is solely determined by \lrd\ (the clustering dimensionality).
This characteristic decouples $m$ and \lrd, which are always equal in previous VQ approaches.
At this time, \lrd\ can vary in a wide range freely.
The changeable \lrd\ contributes to clustering in low-dimensional space, which benefits quantization with lower clustering error.
Besides, it provides an opportunity to comprehensively research the dimensionality reduction and vector quantization without impacting the compressed model size.
Moreover, it introduces a trade-off between the approximation error in low-rank representation learning and the clustering error in quantization.
Based on theoretical analysis and experimental results, we empower the estimation of \lrd\, which appropriately balances errors to achieve better quantization performance.
We experiment \lrvq\ on large-scale datasets like ImageNet and COCO.
Compared with the results in PQF\cite{martinez2021permute}, $\text{LR}^2\text{VQ}$ improves 2.8\%/1.0\% top-1 accuracy for ResNet-18/ResNet-50 on ImageNet with 43$\times$/31$\times$ compression factor, and 0.91\% box AP for Mask R-CNN on COCO with 26$\times$ compression factor.
In summary, our main contributions are as follows:
\begin{itemize}
\item We propose a simple yet effective quantization method called Low-Rank Representation Vector Quantization ($\text{LR}^2\text{VQ}$).
The subvector size $m$ and the clustering dimensionality \lrd\ are mutually independent, making it possible to achieve dimensionality reduction for better quantization performance.
\item We identify the trade-off of \lrd, and provide estimations to \lrd\ with both theoretical and empirical results to balance the approximation error and the clustering error.
\item We evaluate \lrvq\ with various \lrd, which produces state-of-the-art performance on different datasets and architectures.
\end{itemize}

\section{Related Work}
There is a large body of literature on deep neural network compression.
We review related literature from four perspectives: compact network design, punning, tensor decomposition, and quantization.

\subsubsection{Compact network design}
Light weight networks like SqueezeNet\cite{iandola2016squeezenet}, NasNet\cite{zoph2018learning}, ShuffleNet\cite{zhang2017shufflenet}, MobileNets\cite{sandler2018mobilenetv2,howard2019searching} and EfficientNets\cite{tan2019efficientnet} are proposed to be computation and memory effecient.
However, these architectures are either hand-crafted or produced by searching algorithms.
It is inefficient to design networks manually, and current searching approaches demand vast computation resources.
To avoid these issues, another line of work directly operates on the existed network architectures (like VGG or ResNets) to achieve compression and acceleration.

\subsubsection{Pruning}
The simplest pruning operates on filters by removing the connections according to an importance criteria until achieving desirable compression-to-accuracy trade-off\cite{NIPS1989_6c9882bb,guo2016dynamic,han2015learning}.
However, most pruning methods discard individual parameters, and the pruned networks are too sparse to achieve acceleration on embedded devices.
Hence, some approaches focus on pruning unimportant channels to realize compression and acceleration simutanously\cite{he2017channel,li2017pruning,luo2017thinet}.


\subsubsection{Quantization}
Quantization amounts to reducing the bitwidth of each parameter in neural networks.
In this context, we focus on Vector Quantization(VQ), which treats each vector individually to decompose a high-dimensional tensor with small-sized codebooks and low-bit codes.
BGD\cite{stock2020bit}, P\&G\cite{Chen2020TowardsCN}, PQF\cite{martinez2021permute} and DKM\cite{cho2021dkm} are recently proposed vector quantization approaches.
BGD minimizes the reconstruction error of feature maps in the quantized network and optimizes codebooks via layer-wise distillation.
P\&G directly minimizes the reconstruction error of parameters to achieve improvements.
PQF addresses the invariance problem for quantization and formulates a permutation-based method to find a functional equivalent network for easier clustering.
Nevertheless, the permutation is also performed on high-dimensional space, which is more likely to result in a large clustering error.
DKM is an attention-based method that learns to cluster during network training.
However, the differentiable clustering method requires expensive end-to-end training.

Unlike these approaches, our method takes advantage of learning LRRs to achieve changeable clustering dimensionality, which benefits the overall reconstruction error for quantization.
Based on this property, we exploit the trade-off in \lrvq\ to guide the searching for the proper clustering dimensionality.
The accuracy recovery process in \lrvq\ is also high efficiency, which needs a few fine-tuning epochs over the task loss.
All these contributions result in an efficient and effective method for deep neural network compression.

\begin{figure*}[t]
\centering
\includegraphics[width=0.9\textwidth]{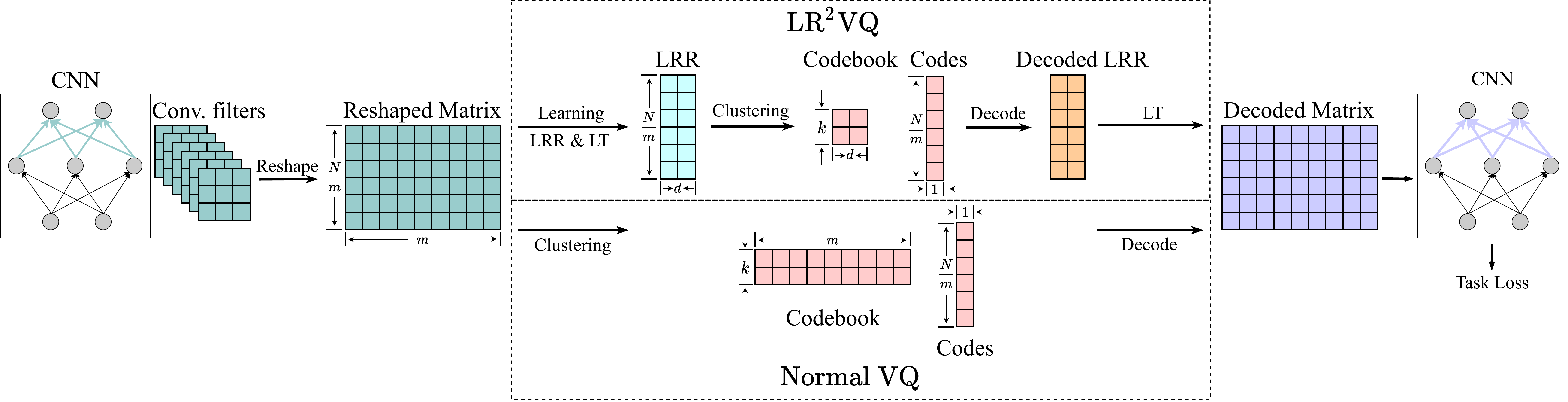}
\caption{The compression pipeline of \lrvq\ and normal VQ. $N$ is the number of parameters in Conv. filters. Unlike normal VQ, \lrvq\ first learns low-rank representation (LRR) and linear transformation (LT), then quantizes LRR with a smaller-sized codebook and low-bit codes. The decoded LRR can be transformed to approximate the reshaped matrix for computation. \lrvq\ can perform clustering on various dimensionality, and the compression ratio is unchanged.}
\label{fig:overall}
\end{figure*}

\section{Method}
Our proposed method aims to achieve a target compression ratio with codebooks and codes from different clustering dimensionality.
Toward this goal, $\text{LR}^2\text{VQ}$ consists of three steps:

\begin{enumerate}
\item \textbf{Learning low-rank representations}:
This step learns low-rank representations (LRRs) for all the convolutional filters using gradient-based learning methods.
We use LRRs and linear transformations (LTs) to replace the \emph{convolutional filters} for computation.
\item \textbf{Quantizing low-rank representations}:
The dimensionality of LRRs is usually different from the original subvectors.
After learning LRRs, we generate codebooks and codes by clustering the subvectors in LRRs.
This operation realizes the variation of clustering dimensionality.
\item \textbf{Global fine-tuning}:
We fine-tune codebooks by minimizing the loss function over training datasets with gradient-based optimization.
After fine-tuning, we merge codebooks and linear transformations (LTs) to eliminate the additional computation complexity during inference.
\end{enumerate}

\subsection{Learning Low-Rank Representations}
\label{sec:lrr}
This section presents how to generate convolutional filters by LRRs and initialize LRRs for robust learning.
\subsubsection{Definition}
Let us denote a convolutional filter $\mathbf{W}\ni \mathbb{R}^{C_{out}\times C_{in}\times K\times K}$ in uncompressed neural networks, with $C_{out}$ as the number of output channels, $C_{in}$ as the number of input channels, and $K$ as the spatial size of the filter.
In normal VQ, \lrw\ should be reshaped into a 2D matrix $\mathbf{W}_r \in \mathbb{R}^{\frac{C_{out}C_{in}KK}{m} \times m}$, where $m$ is the size of subvectors, and $C_{out}C_{in}KK/m$ is the number of subvectors.
The value of $m$ determines the number of subvectors and compression ratios.
In our method, we construct $\mathbf{W}$ by approximating \lrwr\ with two 2D matrices \lra\ and \lrb:
\begin{equation}
\mathbf{W}_r \approx \mathbf{W^{'}} = \mathbf{A} \times \mathbf{B},
\label{eq:mm}
\end{equation}
where $\mathbf{A} \in \mathbb{R}^{\frac{C_{out}C_{in}KK}{m} \times \tilde{d}}$ and $\mathbf{B} \in \mathbb{R}^{\tilde{d}\times m}$.
The dimensionality of $\mathbf{A}$ is $\tilde{d}\in [1,m]$.
Once \lrwp\ is computed, we can reshape it into a 4D tensor of shape $C_{out}\times C_{in}\times K\times K$ to replace \lrw\ for computation.

Equation \ref{eq:mm} is similar to low-rank matrix decomposition, so $\mathbf{A}$ can be treated as the LRR of $\mathbf{W^{'}}$ and $\mathbf{B}$ is the corresponding linear transformation from \lrd\ to $m$.
Instead of mathematical decomposing \lrw, we directly learn \lra\ and \lrb\ by end-to-end optimization, which is more adaptive and efficient for obtaining expressive LRR for \lrwp.

\subsubsection{Initialization}
Weights initialization strongly influences the neural network's optimization and final performance.
As we replace the original weights by $\mathbf{W^{'}}$, the initialization of $\mathbf{W^{'}}$ should be consistent with $\mathbf{W}$.
Thus, we expect $Var(\mathbf{W}) = Var(\mathbf{W^{'}}) = \Delta$.
Based on Equation \ref{eq:mm}, \lrwp\ is computed by the multiplication between \lra\ and \lrb.
For simplicity, we initialize \lra\ with a zero-mean, $\Delta$ variance normal distribution.
With the above assumptions, we only need to compute the variance of \lrb\ for initialization.
According to the derivations in \cite{he2015delving}, we can calculate $Var(\mathbf{B})$ with the following equations:
\begin{equation}
\begin{aligned}
Var(\mathbf{A})\times mVar(\mathbf{B}) &= Var(\mathbf{W}) = Var(\mathbf{W^{'}}) \\
\Rightarrow \Delta \times mVar(\mathbf{B}) &= \Delta\\
\Rightarrow Var(\mathbf{B}) &= \frac{1}{m}.
\end{aligned}
\label{eq:init}
\end{equation}
To preserve the magnitude in the backward pass, we multiply $Var(\mathbf{B})$ with its output dimension $m$.
At this time, we can initialize \lrb\ with a zero-mean, $1/m$ variance normal distribution.

\subsection{Quantizing Low-Rank Representations}
\label{sec:vq}
In this section, we first demonstrate how we apply VQ on the learned LRRs to obtain codebooks \cb \ and codes \cd.
Then, we discuss the trade-off of \lrd\ in \lrvq.
Finally, we introduce an analytic method to search for a coarse estimation of \lrd\ that may result in a lower reconstruction error.

\subsubsection{Codebook and codes}
\label{sec:cbcd}
In \lrvq, LRRs take a majority of network parameters.
So we apply VQ on LRRs to save storage with codebooks \cb \ and codes \cd.
In our definition, \lra \ is matrix with $N/m$ rows and \lrd \ columns, where $N = C_{out}C_{in}KK/$.
We treat each row in \lra \ as an individual subvector to be compressed, so the total number of subvectors for quantization is $N/m$, and the dimensionality of subvectors is \lrd.
To save the storage of these subvectors, we use $k$ centroids with size \lrd \ for approximation.
We call the set of centroids as codebook $\boldsymbol{C}=\{\mathbf{c}_1,...,\mathbf{c}_k\} \in \mathbb{R}^{k\times \tilde{d}}$ where each row in $\boldsymbol{C}$ is a centroid.
Codes $\boldsymbol{I}\in \mathbb{R}^{\frac{N}{m}}$ is a set of assignments that identify the best approximation mapping between subvectors and centroids
\begin{equation}
\boldsymbol{I}_p = \mathop{argmin}\limits_q ||\mathbf{A}_p - \boldsymbol{C}_q||_2^2,
\label{eq:assign}
\end{equation}
where $p$ and $q$ are the row indexes in \lra \ and \cb, and $\mathbf{A}_p\in \mathbb{R}^{1\times \tilde{d}}$ is the $p$th subvector, $\boldsymbol{C}_q \in \mathbb{R}^{1\times \tilde{d}}$ is the $q$th centroid, $\boldsymbol{I}_p\in \mathbb{R}^1$ is a single code to $\mathbf{A}_p$.
With \cb\ and \cd, we can decode them to construct \ah\ by looking up $\boldsymbol{C}$ with all the codes $\boldsymbol{I}$
\begin{equation}
\mathbf{\widehat{A}} =\boldsymbol{C}(\boldsymbol{I})= \{\boldsymbol{C}_{I_1}, ..., \boldsymbol{C}_{I_{\frac{N}{m}}}\} \in \mathbb{R}^{\frac{N}{m} \times \tilde{d}},
\label{eq:lut}
\end{equation}
then transform $\mathbf{\widehat{A}}$ to $m$-dimensional space
\begin{equation}
\mathbf{W}^{'} \approx \mathbf{\widehat{W}^{'}} = \mathbf{\widehat{A}} \times \mathbf{B}.
\label{eq:appmm}
\end{equation}
All the codebooks and codes can be generated by clustering.

The subvector size in normal VQ is $m$, while the subvector size in our method is \lrd.
Note that the clustering dimensionality equals the size of subvectors for all VQ methods.
Therefore, \lrd\ is the clustering dimensionality in \lrvq, which can vary within $[1,m]$.
The number of subvectors in LRR is also directly controlled by $m$, resulting in equivalent number of subvectors to the original filters.
As a result, clustering on LRRs does not impact the compression ratio.
Figure \ref{fig:overall} depicts the comparison between normal VQ and \lrvq.

\lrvq\ is compatible with any clustering methods.
Here, we introduce how \lrvq\ works with \textbf{\textit{k}-means} clustering.
After initializing a codebook with the subvectors in \lra, \textit{k}-means loops within the following steps:
\begin{enumerate}
\item Update assignments via Equation \ref{eq:assign} according to the subvectors in LRRs and codebook;
\item Update codebook (centroids) according to assignments.
\end{enumerate}
Once the loop ends, codes are fixed as each subvector should be replaced by a specific centroid.
Although the Euclidean errors between subvectors and centroids might be tiny, such errors introduce a large gap between \lra\ and \ah.
Therefore, extra network fine-tuning is necessary to compensate for performance loss.

\subsubsection{Trade-off of \lrd}
In \lrvq, \lrd\ is the dimensionality for low-rank approximation and vector quantization.
This value simultaneously influences the approximation error in low-rank representation learning and the clustering error in vector quantization, which affects the overall reconstruction error for model compression.
Here, we discuss the trade-off of \lrd\ in \lrvq.

\begin{figure}[h]
\centering
\includegraphics[width=0.85\columnwidth]{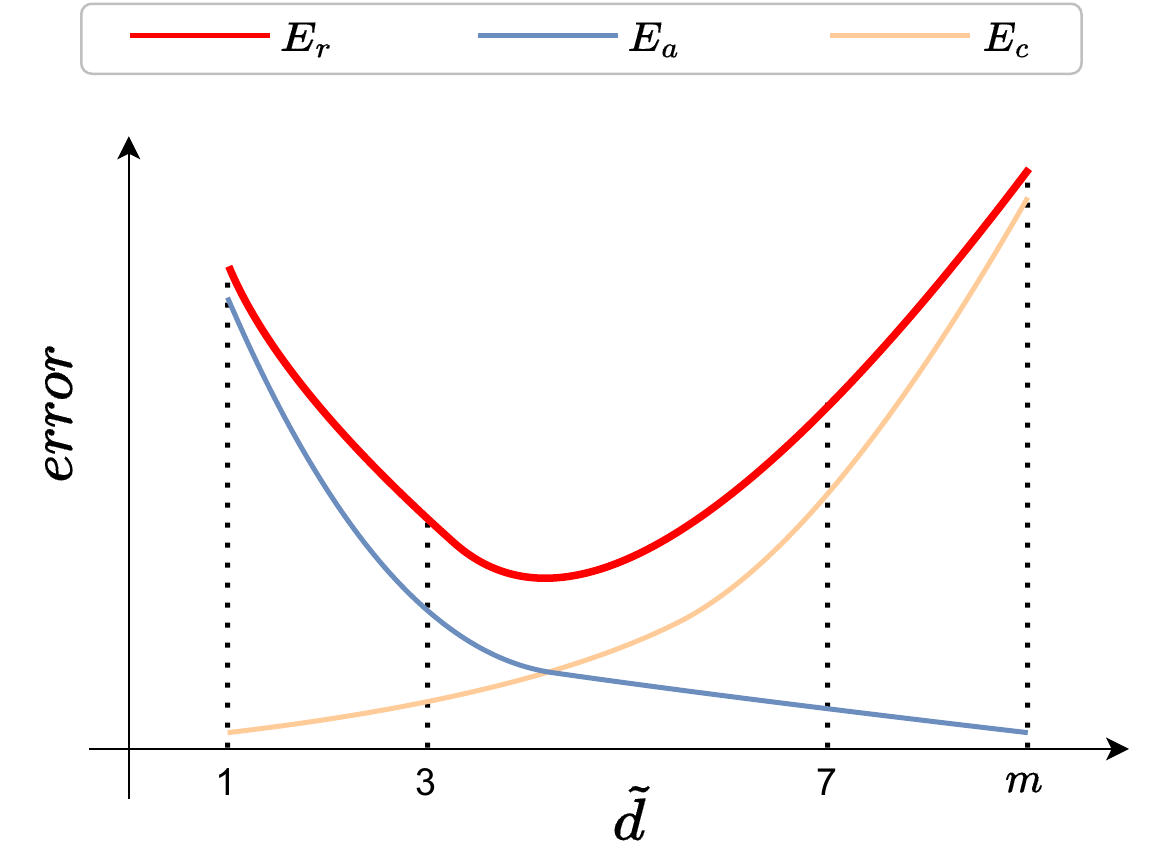}
\caption{Possible variations for errors with $m=9$.}
\label{fig:rec_err}
\end{figure}

The conceptional relationship between the overall reconstruction error $E_r$, the approximation error $E_a$ and the clustering error $E_c$ is described as follows:
\begin{equation}
E_r \propto E_a + E_c,
\end{equation}
and both $E_a$ and $E_c$ are affected by \lrd.
Therefore, we need a proper \lrd\ to reduce these two errors.
For $E_a$, a larger \lrd\ means more parameters in the LRR network, indicating a smaller $E_a$ in \lrvq.
Especially when \lrd\ is close to $m$, the number of parameters in the LRR network is already sufficient to approximate the original network, which can produce a ``zero'' approximation error.
So $E_a$ monotonically decreases with a rising \lrd.
For $E_c$, high-dimensional subvectors are hard to be clustered, resulting in a large clustering error, while the low-dimensional subvectors are more favourable for clustering.
For example, when $\tilde{d}=1$, $E_c$ may become extremely tiny because each subvector is a single scaler in 1D space, which is the lowest and the simplest space for clustering;
When $\tilde{d}=m$, the subvectors in LRR have the same dimensionality as the original subvectors, leading to significant clustering difficulty and the largest clustering error.
So $E_c$ monotonically increases with a rising \lrd.
Based on the above discussion, we can conclude that $E_a$ and $E_c$ are inversely correlated.
Adding these two errors together, we expect that the variation of $E_r$ is similar to Figure \ref{fig:rec_err}, which first declines then increases with a rising \lrd.
Such variation provides reliable insurance that there must be a proper \lrd\ to produce a smaller reconstruction error.
In summary, the trade-off of \lrd\ is the theoretical guarantee in our proposed \lrvq\ to achieve better quantization performance.
We extensively experiment on various \lrd\ in later sections.

\subsubsection{Searching for proper \lrd}
The most effective way to choose proper \lrd\ is grid search, which is time and resources consuming.
Theoretically, an accurate analytical method is useful to search a proper \lrd\ after low-rank representation learning.
Based on the assumption and analysis in PQF\cite{martinez2021permute}, we apply a simple method to coarsely estimate \lrd\ after LRR learning as a starting point.

As different architectures have different characteristics, it is hard to express $E_a$ with mathematics.
Besides, $E_a$ is fixed after LRR learning and can be measured by the model performance.
Consequently, we choose the LRR networks comparable to the original network for searching \lrd\ because $E_a$ in these networks is negligible.
At this time, $E_r$ is dominated by $E_c$, and we only need to consider $E_c$ in searching proper \lrd.
We note that $E_c$ indicates the clustering error between $\mathbf{W}^{'}$ and $\mathbf{\widehat{W}}^{'}$ rather than \lra\ and \ah.
The essence of \lrvq\ is clustering the subvectors in \lrwp, and we achieve this goal with an indirect method, which is clustering on \lra.
Therefore, the clustering error between \lrwp\ and \lrwhp\ is the actual $E_c$ in \lrvq.
Based on the above analysis, the following computations are all performed on \lrwp\ rather than \lra.

According to PQF, we can infer that $\mathbf{W}^{'}\sim \mathcal{N}(\mathbf{0}, \mathbf{\mathbf{\Sigma}})$, where $\mathbf{\mathbf{\Sigma}}\in \mathbb{R}^{m\times m}$ is the covariance of $\mathbf{W}^{'}$, and the lower bound of $E_c$ is
\begin{equation}
E_c \geq k^{-\frac{2}{m}}m|\mathbf{\Sigma}|^{\frac{1}{m}}.
\label{eq:bd}
\end{equation}
Similar to PQF, we minimize $|\mathbf{\Sigma}|$ to reduce $E_c$.

After LRR learning, each LRR network with a candidate \lrd\ adopts the following computations:
\begin{enumerate}
\item Generating \lrwp\ for all convolutional layers;
\item Calculating $|\mathbf{\Sigma}|$ for \lrwp;
\item Adding all the $|\mathbf{\Sigma}|$ to represent $E_c$.
\end{enumerate}
The quantization regimes for all FC layers are the same, and we do not apply \lrvq\ to them.
So the $|\mathbf{\Sigma}|$ in FC layers can be treated as equal, which can be neglected in each network.
After the above computation, the network with the lowest $|\mathbf{\Sigma}|$ provides a coarse estimation for \lrd, and quantizing the corresponding LRR network is more likely to achieve a better compression-to-accuracy ratio in \lrvq.
We demonstrate the results of $|\mathbf{\Sigma}|$ with different \lrd\ in the later section to validate the coarse estimations of our method.

\subsection{Global Fine-tuning}
After clustering, global fine-tuning is necessary to compensate for performance loss caused by $E_c$.
In this step, we fix the assigned centroids to all subvectors and fine-tune the compressed network with the loss function and training data in low-rank representation learning.
During fine-tuning, all the subvectors will be replaced by Equation \ref{eq:lut} and \ref{eq:appmm} for computation.
So the centroids are differentiable and can be updated by gradients as follows:
\begin{equation}
\mathbf{c}(i)_{t+1}^l \leftarrow \mathbf{c}(i)_t^l- \hat{\eta} \frac{\partial{\hat{\mathcal{L}}}}{\partial{\mathbf{c}(i)_t}},
\end{equation}
where $\hat{\mathcal{L}}$ is the loss function, and the learning rate is modified to $\hat{\eta}$.
This procedure is potent in boosting the model performance of the quantized networks.

\subsection{Inference without \lrb}
The linear transformation (LT) \lrb\ is necessary for low-rank representation learning.
Nevertheless, once all the parameters are fixed after global fine-tuning, we can eliminate \lrb\ and the computation in Equation \ref{eq:mm}.
Such elimination is accomplished by the commutative property between LT and Look-Up Table (LUT) operation.
After the codebook \cb\ is fixed, we can obtain a new codebook $\boldsymbol{C}^{'}$ by
\begin{equation}
\boldsymbol{C^{'}} = \boldsymbol{C} \times \mathbf{B} \in \mathbb{R}^{k\times m},
\end{equation}
and the computation of \lrvq\ is modified to
\begin{equation}
\begin{split}
\mathbf{W} \approx \mathbf{W}^{'} \approx \mathbf{\widehat{W}^{'}} &=\boldsymbol{C}^{'}(\boldsymbol{I})\\
& = (\boldsymbol{C}\times \mathbf{B})(\boldsymbol{I})
= \boldsymbol{C}(\boldsymbol{I})\times \mathbf{B}\\ &= \mathbf{\widehat{A}} \times \mathbf{B} .
\end{split}
\label{eq:inf}
\end{equation}

Equation \ref{eq:inf} implies that only codes \cd\ and the new codebook $\boldsymbol{C}^{'}$ are expected to be stored, and the computation in Equation \ref{eq:mm} is completely removed by a simple LUT operation $\boldsymbol{C}^{'}(\boldsymbol{I})$ during inference.

\begin{table*}[h]
\centering
\tiny
\caption{ImageNet results for PQF, vannila DKM and $\text{LR}^2\text{VQ}$. The value of $\tilde{d}_{cv}$ and $\tilde{d}_{pw}$ are shown after \lrvq\ 's results.}
\begin{adjustbox}{width=0.85\textwidth, center}
\begin{tabular}{cccccccc}
\toprule
\multirow{2}{*}{Model}	&\multirow{2}{*}{\shortstack{Original\\top-1}} 	&\multirow{2}{*}{\shortstack{Original\\size}}   &\multirow{2}{*}{\shortstack{Comp.\\size}} &\multirow{2}{*}{\shortstack{Comp.\\ratio}} &\multirow{2}{*}{PQF} &\multirow{2}{*}{DKM} &\multirow{2}{*}{\lrvq\ ($\tilde{d}_{cv}/\tilde{d}_{pw}$)} \\
&&&&&&& \\
\midrule
\multirow{2}{*}{ResNet-18} &\multirow{2}{*}{71.30}  &\multirow{2}{*}{44.59M}    &1.54M &29x &67.98 &68.09 &\textbf{69.56}(4/4)\\
&&&1.03M &43x &64.39 &64.33 &\textbf{67.21}(4/4)\\
\midrule
\multirow{2}{*}{ResNet-50} &\multirow{2}{*}{77.75}	&\multirow{2}{*}{97.49M}   &5.09M &19x &74.93 &oom &\textbf{76.17}(5/4)\\
&&&3.19M &31x &73.40 &73.88 &\textbf{74.48}(5/4)\\
\bottomrule
\end{tabular}
\end{adjustbox}
\label{tb:sota}
\end{table*}

\section{Experiments}

\subsection{Experiments on ImageNet}
In this section, we evaluate our $\text{LR}^2\text{VQ}$ with vanilla ResNet-18, ResNet-50\cite{he2015deep}
on ImageNet\cite{russakovsky2015imagenet} classification datasets.

\subsubsection{Baselines}
We compare our method with PQF and vanilla DKM as they are the most competitive VQ methods up to the writing of this paper.
As $\text{LR}^2\text{VQ}$ requires low-rank pre-training before quantization, we do not use the pre-trained models from the Pytorch model zoo.
Instead, we implement our training procedures and re-implement PQF and vanilla DKM under the same code base for a fair comparison.
As a representative VQ method, PQF has conducted sufficient experiments and set solid baselines, so we follow its compression settings to compare different methods under identical model sizes.

\begin{table}[t]
\large
\caption{Compression regimes with $k_{cv}=k_{pw}=256$.}
\begin{adjustbox}{width=\columnwidth, center}
\begin{tabular}{cccccc}
\toprule
Architecture &Regime &$m_{cv}$ &$m_{pw}$ &$\tilde{d}_{cv}$ &$\tilde{d}_{pw}$ \\
\midrule
\multirow{2}{*}{ResNet-18}	&Small blocks &9 &4 &\multirow{2}{*}{$[1,m_{cv}]$}	&\multirow{2}{*}{$[1,m_{pw}]$}\\

&Large blocks &18 &4 & &\\
\midrule
\multirow{2}{*}{ResNet-50}	&Small blocks &9 &4 &\multirow{2}{*}{$[1,m_{cv}]$} &\multirow{2}{*}{$[1,m_{pw}]$}\\
&Large blocks &18 &8 &&\\
\bottomrule
\end{tabular}
\end{adjustbox}
\label{tb:config}
\end{table}

\subsubsection{Compression setups}
Let's denote $cv$ for $3\times3$ convolution, $pw$ for $1\times 1$ convolution and $fc$ for fully-connected layer.
To identify the hyperparameters in different convolutional layers in \lrvq, we use $m_{cv}, m_{pw}, k_{cv}, k_{pw}, \tilde{d}_{cv}, \tilde{d}_{pw}$ to represent $m$, $k$ and \lrd \ in $3\times3$ and pointwise convolution.
We set two compression regimes to achieve different compression ratios.
The detailed configurations are shown in Table \ref{tb:config}.
The large blocks regime means fewer codes and a smaller model size for quantized networks.
Specifically, the clustering dimensionality for $3\times3$ and pointwise convolution in PQF and vanilla DKM are equal to $m_{cv}, m_{pw}$, and $k_{cv}$, $k_{pw}$ are also the same as in \lrvq.
For FC layers, the dimensionality of subvectors is 4, and $k=2048$ for ResNet-18 and $k=1024$ for ResNet-50.
For other settings, we follow \cite{martinez2021permute} that we clamp the number of centroids to $min(k, N/4)$ for stability, and we do not compress the first $7\times 7$ convolution in ResNets since they take less than 0.1\% model size.

\subsubsection{Memory footprint}
Following PQF, we only compress the weights in convolutional and FC layers and ignore the bias in FC and batchnorm layers.
We train networks with 32-bit floats but store the codebooks with 16-bit floats.
For $k=256$, all the codes can be stored as 8-bit integers.
These settings effectively reduce model sizes with negligible accuracy loss.

\subsubsection{Training details}
The low-rank representation learning uses a total batch size of 1024 on 16 NVIDIA V100 GPUS to train 100 epochs with SGD optimizer plus Nesterov and momentum 0.9.
Weight decay is 0.0001, and the learning rate is 0.4 with a cosine annealing scheduler.
Label smooth is set to 0.1.
We train LRR networks with $\tilde{d}\in[3,7]$ because other values produces either larger $E_a$ or larger $E_c$ to harm quantization.
After LRR learning, we run our searching method to estimate a coarse \lrd, and start quantizing the corresponding LRR network.
We run \textit{k}-means for 100 iterations to obtain codebooks and codes, then fine-tune the compressed network with an Adam optimizer.
The initial learning rate of fine-tuning is 0.001 and annealed by a cosine scheduler.
This procedure runs on 16 GPUs with batch size 2048 for 9 epochs, which takes half an hour for ResNets.

\subsubsection{Results}
Table \ref{tb:sota} shows the comparison of our $\text{LR}^2\text{VQ}$ with \textit{k}-means clustering against PQF and vanilla DKM on standard ResNet-18 and ResNet-50 with the configurations in Table \ref{tb:config}.
The table shows that \lrvq\ outperforms PQF and vanilla DKM across all configurations.
For ResNet-18 with large block compression, \lrvq\ presents a definite 2.8\% improvement over PQF under 43$\times$ compression factor.
For ResNet-50, \lrvq\ consistently outperforms baselines for more than 1\% top-1 accuracy under 31$\times$ compression factor.
Specifically, we mark the proper $\tilde{d}_{cv}$ and $\tilde{d}_{pw}$ after the results of \lrvq.
As can be seen, $d_{cv}$ is 4 or 5 in both compression regimes, which is lower than $d_{cv}=9 \text{ or } 18$ in PQF and vanilla DKM.
This result confirms that low-dimensional clustering effectively reduces the clustering error and the reconstruction error.
Besides, it proves the advances in jointly considering the compression ratio and the clustering dimensionality, which possess great potential in benefiting vector quantization.
Another result we want to discuss here is the performance of vanilla DKM.
Based on the computation of DKM, we expect it to be a better clustering method compared to \textit{k}-means as the clusters are optimized end-to-end.
Contrary to expectations, vanilla DKM has a similar performance to PQF.
This phenomenon may be explained by the optimization of DKM also suffers from high-dimensional clustering, which suggests varying the dimensionality in different clustering methods.

\subsection{Experiments on COCO}
To generalize \lrvq\ to different datasets and architectures, we compress Mask R-CNN with \lrvq\ and experiment on COCO datasets.
We first propose low-rank representation learning from scratch to obtain LRRs, then clustering the subvectors in LRRs with \textit{k}-means and fine-tuning the overall network on COCO datasets.
The pre-trained network size is different from \cite{martinez2021permute} because we utilize the architecture in Detectron2\cite{wu2019detectron2} instead of Pytorch\cite{NEURIPS2019_9015}.
For a fair comparison, the compression configurations for PQF and \lrvq\ are the same as in \cite{martinez2021permute}.
The results are demonstrated in Table \ref{tb:coco}.
Our method obtains 38.20 box AP, which remarkably surpasses PQF with 0.91 AP under 26$\times$ compression factor.
These results illustrate the generalizability of \lrvq\ to different vision tasks and architectures.

\begin{table}[h]
\small
\centering
\caption{Compression results for Mask R-CNN on COCO.}
\begin{tabular}{lcccc}
\toprule
Method &Size &Ratio &Box AP &Mask AP	\\
\midrule
Uncompressed &174.37M &- &40.34 &36.54 \\
PQF &6.65M &26$\times$	&37.29 &34.24 \\
\lrvq &6.65M &26$\times$ &\textbf{38.20} &\textbf{34.93} \\
\bottomrule
\end{tabular}
\label{tb:coco}
\end{table}



\subsection{Ablation on the trade-off of \lrd}
\label{sec:trend}
We have discussed the trade-off of \lrd\ and proposed assumptions on the variation of errors in Figure \ref{fig:rec_err}.
Here, we extensively experiment on \lrd\ to investigate this trade-off and prove our assumption.
We experiment on ImageNet to compare the model performance with different \lrd.
We iterate $\tilde{d}_{cv}$ among $[1,9]$ (or $[1,18]$) with small (or large) block regime in ResNet-18, $\tilde{d}_{pw}$ among $[1,8]$ with large block regime in ResNet-50, and $\tilde{d}_{cv}$ among $[1,9]$ with small block regime in Mask R-CNN.
These settings ensure that the dimensionality of LRRs varies in a wide range.
Other configurations are the same as in Table \ref{tb:config}.
The baseline we compared here is PQF.
We plot the accuracy of low-rank pre-trained networks and their corresponding quantized networks.
Surprisingly, these curves demonstrate a similar trend across different configurations and architectures.
As \lrd\ rises, the accuracy of LRR pre-trained models rapidly increases, then oscillates around the original uncompressed network.
The quantized networks' curves also perform similarly across different settings, which first increase to the top and then decline with a rising \lrd.
These tendency broadly support the assumptions on $E_a$ and $E_r$ in Figure \ref{fig:rec_err} and further guarantees the proper \lrd\ can appropriately balance $E_a$ and $E_c$ to achieve lower $E_r$.
All the peaks of red curves are the empirical results to the proper \lrd\ in \lrvq.
Specifically, the accuracy of compressed and uncompressed ResNet-18 with $\tilde{d}=1$ is extremely tiny, indicating that the clustering error in 1D space is negligible, so $E_a$ dominates $E_r$ to produce a poor performance for the compressed and uncompressed networks.
Almost all the LRR pre-trained networks become comparable to the original networks before reaching $\tilde{d}=m$, which implies enormous parameter redundancy in convolutions.
We note that $\tilde{d}_{cv}$ is fixed to 5 in ResNet-50 with large blocks, which limits the power of $3\times3$ convolutions.
So there is always a gap between LRR pre-trained networks and the original uncompressed network.
Taking all these results together, the trade-off of \lrd\ provides reliable insurance that there must be a proper \lrd\ in \lrvq\ to achieve better quantization performance.

\begin{figure}[h]
\centering
\begin{subfigure}{\columnwidth}
\centering
\includegraphics[width=\textwidth]{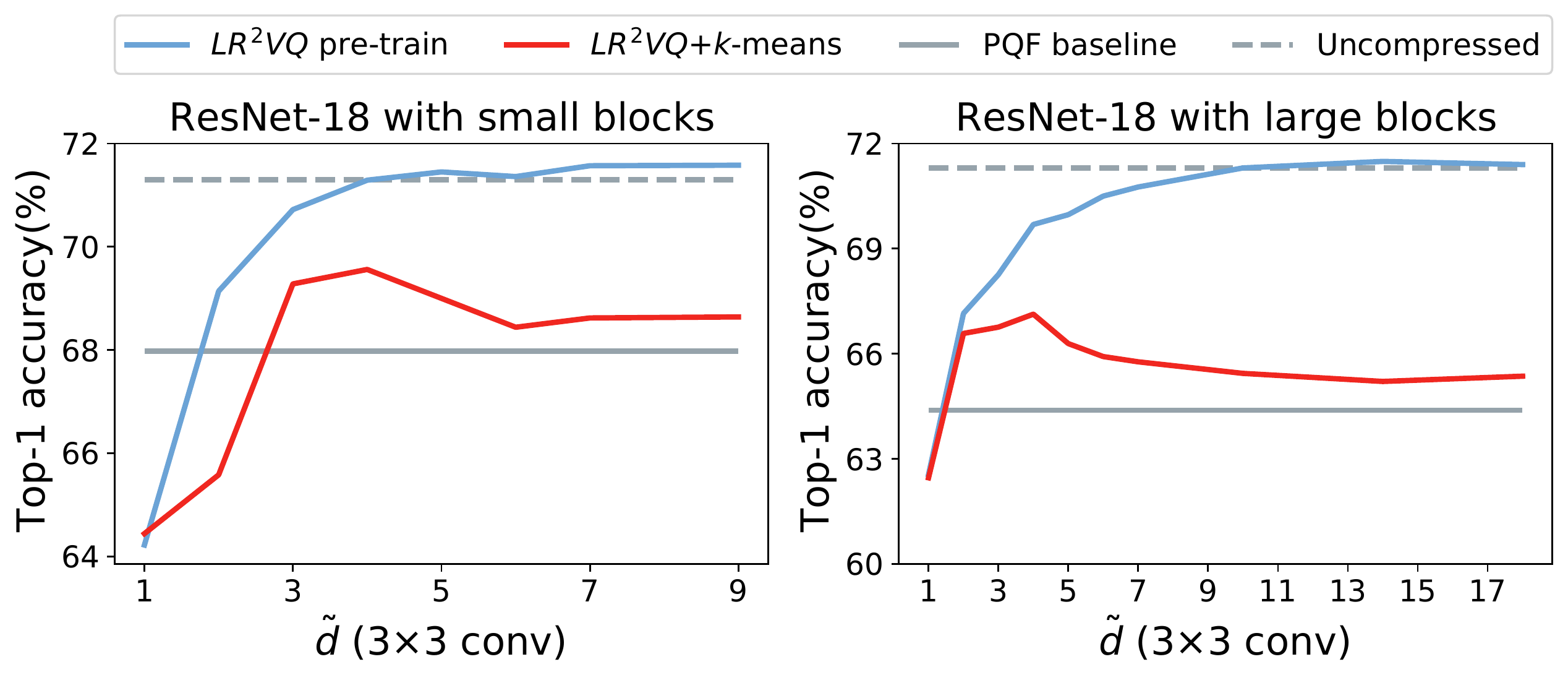}
\end{subfigure}
\begin{subfigure}{\columnwidth}
\centering
\includegraphics[width=\textwidth]{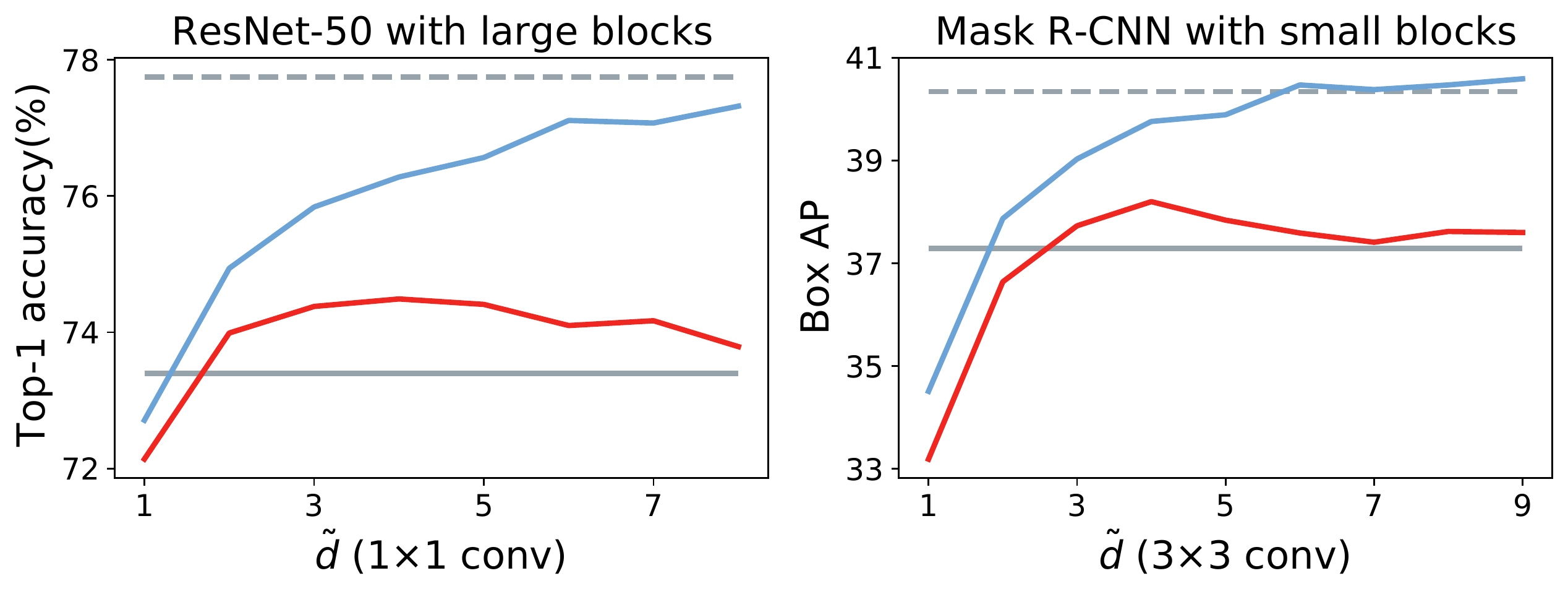}
\end{subfigure}
\caption{Top-1 accuracy with different dimensionality.}
\label{fig:abl-d}
\end{figure}

\subsubsection{Lower intrinsic dimensionality}
\begin{figure}[h]
\centering
\includegraphics[width=\columnwidth]{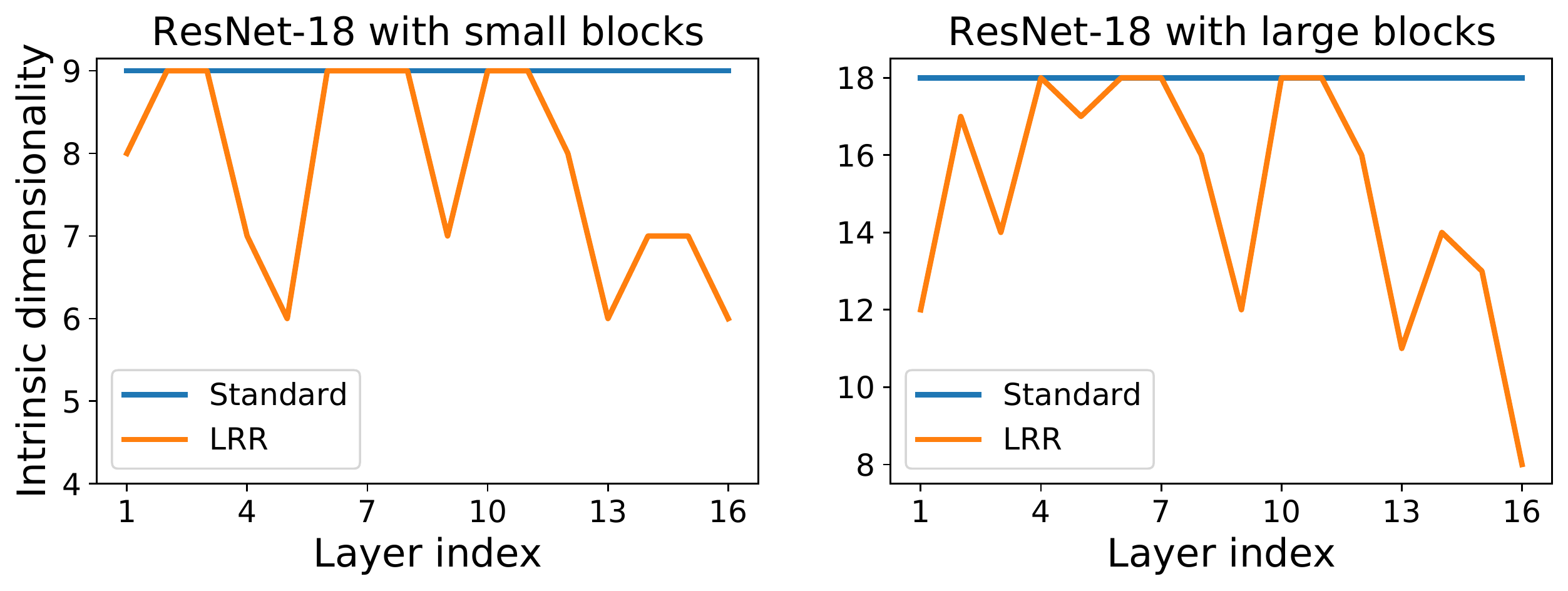}
\caption{Intrinsic dimension of subvectors before clustering. \textbf{Standard} and \textbf{LRR} means pre-raining with vanilla filter and low-rank representation filter.}
\label{fig:indim}
\end{figure}

One unanticipated finding in Figure \ref{fig:abl-d} is that \lrvq\ outperforms PQF even at $\tilde{d}_{cv}=m_{cv}$.
Intuitively, the clustering difficulty for $\tilde{d}=m$ is similar to PQF, so the performance of \lrvq\ and PQF should be close.
However, in such circumstances, \lrvq\ outperforms PQF in all architectures.
This discrepancy may be attributed to our low-rank representation learning, which implicitly learns lower intrinsic dimensionality to make the subvectors more favourable for clustering.
To validate this speculation, we plot the intrinsic dimensionality of LRRs by principal components analysis (PCA) in each layer with $\tilde{d}_{cv} = m_{cv}$ in Figure \ref{fig:indim}.
The x-axis is the layer index, and the y-axis is the intrinsic dimensionality with a variance ratio of more than 99.99\% after PCA.
As the figure shows, the intrinsic dimensionality of LRRs tends to be much lower than the original filters.
For example, 11 layers learn lower intrinsic dimensionality among 16 layers in large block compression.
These results suggest that LRRs can automatically learn lower intrinsic dimensionality to benefit clustering.

\subsection{Results of searching \lrd}
\begin{figure}[h]
\centering
\includegraphics[width=\columnwidth]{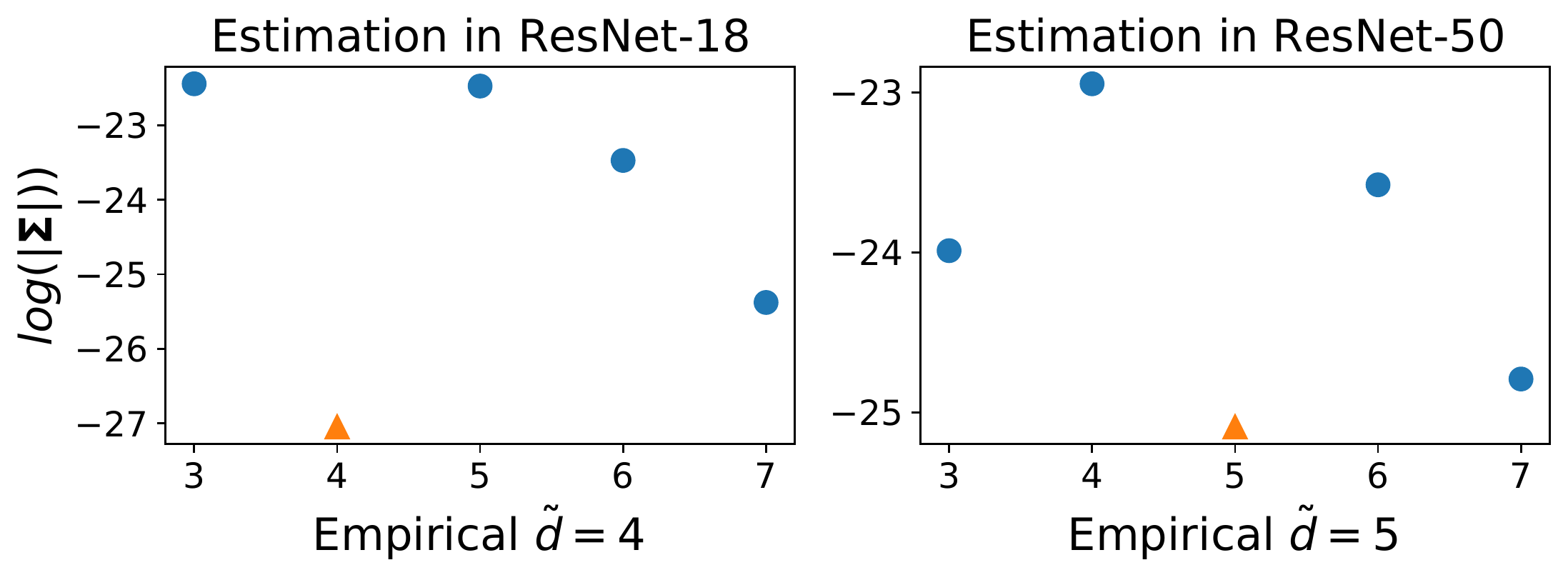}
\caption{The estimation of our coarse analytical method.
Empirical means \lrd\ is given by the experimental results from Figure \ref{fig:abl-d}.
\lrd\ with a lower $|\mathbf{\Sigma}|$ is the better estimation.}
\label{fig:sd}
\end{figure}

To reduce the fine-tune overhead, a theoretical analysis to refine candidate \lrd s is very valuable.
Here, we demonstrate the estimation of our coarse method for searching \lrd.
As described in previous sections, we apply our searching method with $\tilde{d}\in [3,7]$.
Figure \ref{fig:sd} presents the summation of $|\mathbf{\Sigma}|$ for different \lrd.
As a starting point, our coarse estimation of the proper \lrd\ using Equation \ref{eq:bd} shows agreements with experimental results.
All these results are reasonable because the distribution of subvectors also affects clustering, which is consistent with the results in the previous section.
With a fixed number of centroids, reducing \lrd\ can decrease the clustering difficulty, but hard-for clustering subvectors also produce significant clustering errors.
Fortunately, our experiments show that the learned subvectors in \lrvq\ are favourable for clustering.
Hence, our coarse estimation provides the right direction towards the proper \lrd.

\section{Conclusion}
We propose a new method called \lrvq, which first learns low-rank representations (LRRs) and then quantizes LRRs to achieve compression.
\lrvq\ decouples the subvector size and the clustering dimensionality by quantizing the subvectors in the learned LRRs, making it possible to realize the variation of clustering dimensionality under a fixed compressed model size.
The changeable nature of clustering dimensionality introduces a trade-off between the approximation error and the clustering error, which implies that the value of \lrd\ is critical to the performance of \lrvq.
We provide theoretical analysis and empirical observations to offer estimations of the proper \lrd\ after LRR learning.
We evaluate \lrvq\ on different datasets and architectures, and all the results demonstrate that \lrvq\ leads the state-of-the-art performance among competitors.
This paper provides the first comprehensive assessment of reducing clustering dimensionality, which is trustworthy for vector quantization.

\bibliographystyle{IEEEtran}
\bibliography{seconv_ref}

\end{document}